\documentclass[conference]{IEEEtran}
\IEEEoverridecommandlockouts

\usepackage{cite}
\usepackage{amsmath,amssymb,amsfonts}
\usepackage{algorithmic}
\usepackage{subcaption}
\usepackage{graphicx}
\usepackage{textcomp}
\usepackage{xcolor}
\usepackage{booktabs}
\usepackage{caption}
\title{Riverbank Erosion Analysis in Bangladesh Using Spatiotemporal Segmentation}

\author{
\IEEEauthorblockN{
M. Saifuzzaman Rafat\IEEEauthorrefmark{1},
Akif Islam\IEEEauthorrefmark{1},
Mohd Ruhul Ameen\IEEEauthorrefmark{1},\\
Momen Khandoker Ope\IEEEauthorrefmark{1},
Abu Saleh Musa Miah\IEEEauthorrefmark{2},
Jungpil Shin\IEEEauthorrefmark{2}
}
\IEEEauthorblockA{\IEEEauthorrefmark{1}University of Rajshahi, Rajshahi, Bangladesh\\
Email: \{saif.rafat, iamakifislam, ameensunny242\}@gmail.com, khandokermomen919@ru.ac.bd}
\IEEEauthorblockA{\IEEEauthorrefmark{2}University of Aizu, Aizuwakamatsu, Japan\\
Email: \{musa, jpshin\}@u-aizu.ac.jp}
}

\begin{document}


\maketitle

\begin{abstract}
Riverbank erosion is a serious environmental problem in Bangladesh, causing land loss, damage to infrastructure, and displacement of local communities. Manual analysis of satellite images is often slow and difficult to apply consistently across large river networks. This study uses a parameter-efficient adaptation of the Segment Anything Model (SAM) to detect and measure riverbank erosion from historical Google Earth images. A dataset of 500 image pairs from 2003 to 2025 was prepared from erosion-prone areas, including Mokterer Char, Kedarpur, and Chowhali Upazila, with pixel-level labels for river, stable land, and eroded regions. During training, the ViT-H image encoder and prompt encoder were kept frozen, while only the lightweight mask decoder was fine-tuned for riverine segmentation. The adapted model achieved an erosion-class IoU of 0.867 and an F1-score of 0.928 on the primary held-out test set. Evaluation on unseen riverbank regions also showed that the model could generalize to new geographic areas, although detecting accreted land from RGB-only images remained difficult. The estimated erosion area differed from the ground truth by only 0.17\%, showing that the model can produce reliable area measurements. Overall, this study demonstrates that adapted foundation segmentation models can support faster and more consistent riverbank erosion monitoring, with future scope for using multi-modal remote sensing data in broader environmental assessment.
\end{abstract}


\section{Introduction}
Bangladesh's economy, landscape, and daily life are closely connected to the Padma–Brahmaputra–Meghna river system. These rivers nourish agriculture and commerce, yet their shifting courses consume thousands of hectares annually. An estimated 10,000 hectares vanish each year into the current \cite{nwmp2001}, and since 1967 the Padma River alone has swallowed more than 66,000 hectares of land \cite{nasa2018padma}. Between 1973 and 2004, the Jamuna erased nearly 878 square kilometers while the Padma claimed another 294 \cite{cegis2009}. These figures confirm that riverbank erosion is a serious and continuing problem in Bangladesh.

The impacts extend beyond physical land loss. Erosion destroys homes, roads, farmland, and entire settlements, displacing communities that may have lived in place for generations. In 2018, the Padma engulfed Mokterer Char and Kedarpur, displacing thousands \cite{newage2018}. In 2017, the Brahmaputra swept away Balchipara, leaving around one hundred families homeless \cite{reuters2017}. Such incidents reflect a recurring pattern caused by the unstable, dynamic nature of Bangladesh's river systems, making effective monitoring essential for identifying vulnerable areas before damage becomes more severe.

Traditional monitoring approaches are no longer sufficient at this scale. Field surveys and manual interpretation of satellite imagery are slow, expensive, and labor-intensive, and their usefulness is limited by seasonal variation and cloud cover \cite{kar2014,freihardt2023}. With an area of about 148,000 km$^2$ and a river network of approximately 24,140 km \cite{bbs2011,islam2011}, these methods cannot be applied continuously across the whole country. An automated approach is therefore needed for faster, more efficient detection of riverbank change.

Recent advances in artificial intelligence offer a promising direction. The Segment Anything Model (SAM), introduced by Meta AI, has shown strong performance in image segmentation and other visual tasks \cite{kirillov2023sam}. In remote sensing, SAM has already been applied to waterbody mapping \cite{rana2023}, coastal boundary detection \cite{blais2025}, and agricultural field segmentation \cite{huang2024}. However, its application to riverbank erosion remains limited. This is a difficult task because river water, wet sediment, and sandy banks often appear visually similar, while channels change shape over time, making accurate segmentation challenging.

Previous studies used remote sensing and GIS techniques to analyze riverbank erosion in Bangladesh \cite{hassan2017,rahman2017}. Although useful for documenting past changes, these methods are not suitable for large-scale, detailed monitoring without automation. Another major limitation is the lack of publicly available pixel-level data on riverbank erosion and settlement loss in Bangladesh.

This paper addresses these gaps by combining adapted foundation-model segmentation with temporal change analysis for automated riverbank erosion monitoring. The main contributions of this work are:
\begin{enumerate}
    \item A manually annotated dataset of 500 historical Google Earth image pairs (2003--2025) from the Padma and Jamuna rivers with pixel-level labels for river, stable land, and eroded zones.
    \item A parameter-efficient fine-tuning strategy for SAM that freezes the ViT-H encoders while updating only the 4-million-parameter mask decoder.
    \item An automated temporal-differencing pipeline for quantitative erosion and accretion estimation in km$^2$.
    \item Validation showing an erosion-class IoU of 0.867, F1-score of 0.928, and erosion area error of 0.17\% on the geographically unseen evaluation set, with an analysis of RGB-only accretion challenges.
\end{enumerate}

\begin{figure*}[!t]
  \centering
  \includegraphics[width=\textwidth,keepaspectratio]{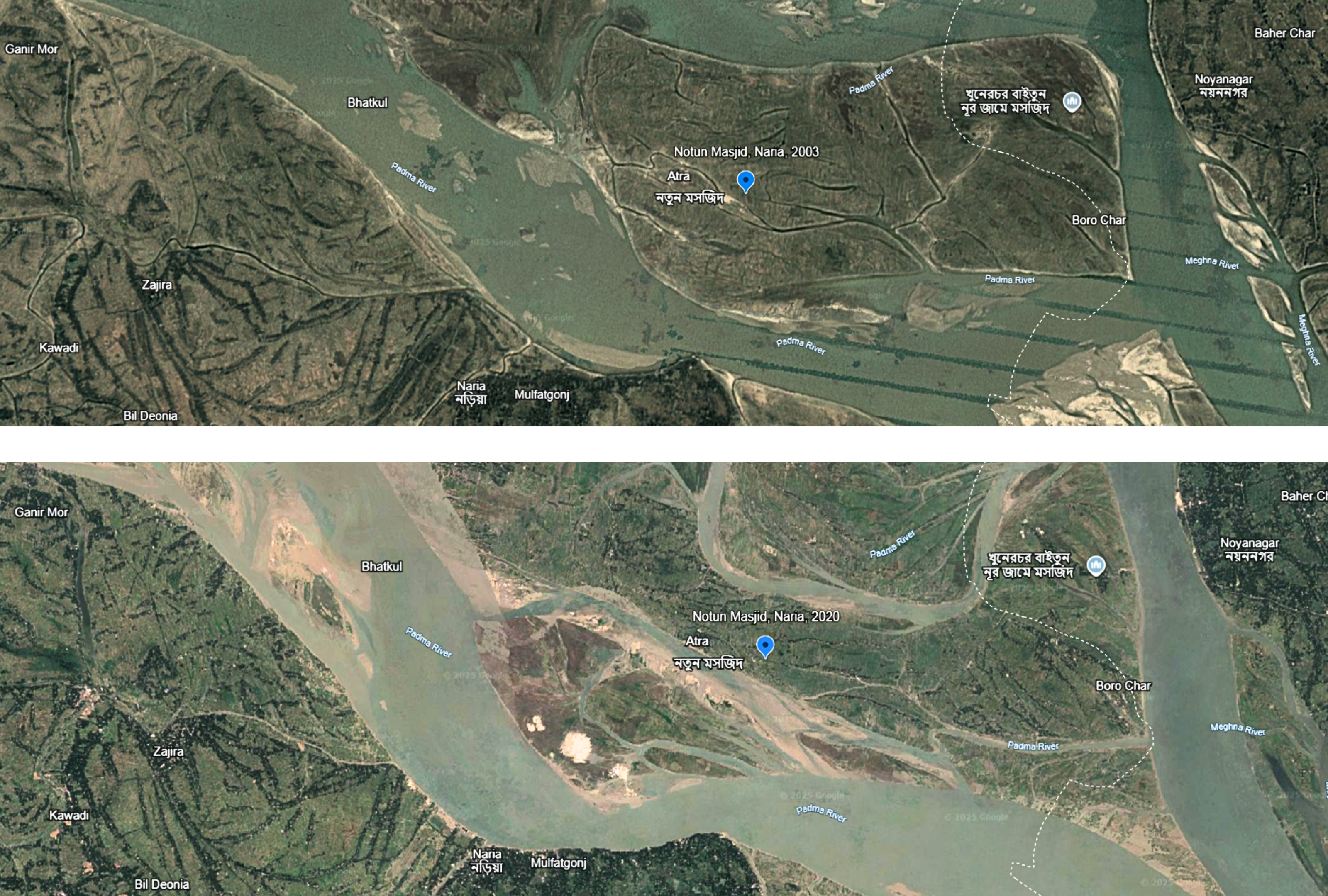}
  \caption{Satellite imagery of Naria Upazila showing riverbank changes between 2003 (top) and 2020 (bottom). The marker indicates the site of Notun Masjid, beside which a drastic loss of land area has occurred due to river erosion. The comparison reveals significant channel migration and reduction of landmass over time.}
  \label{fig:naria_decade_comparison}
\end{figure*}

\section{Related Works}

\subsection{Traditional and ML-Based Approaches}
Morphological analysis of river dynamics has traditionally relied on manual digitization of satellite imagery and Geographic Information Systems (GIS). Studies by Bhuiyan et al. \cite{bhuiyan2017} and Rahman et al. \cite{rahman2017} quantified the Padma River's erosion, estimating losses of over 189 km² between 1973 and 2011. Similarly, Hassan et al. \cite{hassan2017} utilized satellite data to track land loss along the Jamuna River. While foundational, these manual methods are labor-intensive, subject to inter-annotator bias \cite{chu2006, islam2009erosion}, and constrained by the availability of cloud-free optical data in monsoon climates \cite{kummu2008}.

To address these limitations, recent works have incorporated radar data (SAR) for all-weather monitoring \cite{freihardt2023} and machine learning techniques such as fuzzy logic \cite{alam2023} and Artificial Neural Networks (ANNs) \cite{sarker2024} for risk prediction. However, these methods often rely on handcrafted features or point-based data and struggle to capture the pixel-level spatial details required for precise settlement loss quantification.

\subsection{Deep Learning in Remote Sensing}
The advent of Convolutional Neural Networks (CNNs), particularly the U-Net architecture, has significantly advanced semantic segmentation in remote sensing. Nguyen et al. \cite{nguyen2022} and Liu et al. \cite{liu2024} demonstrated high accuracy in coastline and waterbody extraction, while Boonpook et al. \cite{boonpook2018} achieved over 90\% accuracy in building extraction. Despite their success, standard CNNs face two critical challenges in riverine contexts: (1) they require large volumes of task-specific labeled data, which is non-existent for historic river erosion in Bangladesh; and (2) they often struggle with domain shifts, particularly the spectral confusion between sediment-rich water and riverbanks.

\subsection{Foundation Models and SAM}
The Segment Anything Model (SAM) \cite{kirillov2023sam} represents a paradigm shift toward foundation models capable of zero-shot generalization. Leveraging a vision transformer trained on 1 billion masks, SAM has outperformed traditional supervised models in various remote sensing tasks, including aquaculture mapping \cite{li2023} and agricultural field delineation \cite{huang2024}. Recent studies have further improved SAM's efficiency through parameter-efficient fine-tuning, adapting the model with minimal computational overhead \cite{paranjape2024, xie2024}. 

However, the application of SAM to \textit{riverbank erosion} remains largely unexplored. Existing studies primarily focus on static object detection; they do not address the challenge of temporal change detection in highly dynamic, sediment-rich environments. This study addresses this gap by fine-tuning SAM on a temporal riverbank erosion dataset and applying it to automated land-loss quantification.

\section{Methodology}

\subsection{Dataset Curation and Preprocessing}

\begin{table}[!t]
\caption{Riverbank Erosion Dataset Specifications}
\label{tab:erosion_dataset}
\centering
\footnotesize
\begin{tabular}{ll}
\toprule
\textbf{Parameter} & \textbf{Specification} \\
\midrule
Total Samples & 500 image pairs (pre/post erosion) \\
Source & Google Earth Pro (Maxar, CNES/Airbus, Landsat) \\
Temporal Scope & 2003--2025 \\
Spectral Channels & 3 (Red, Green, Blue) \\
Native Resolution & 0.5 m -- 15 m (varies by sensor) \\
Target Resolution & Resampled to 10 m/pixel \\
Image Dimensions & $1024 \times 1024$ pixels \\
Cloud Cover & $<10\%$ (filtered for Dry Season: Nov--Mar) \\
Split & Train (50\%) / Val (10\%) / Test (40\%) \\
\bottomrule
\end{tabular}
\end{table}

A comprehensive dataset was curated from erosion-prone river basins across Bangladesh, specifically targeting the Padma and Jamuna riverbanks. Imagery was sourced via Google Earth Pro, integrating historical archives from providers such as Maxar, SPOT, and Landsat. To mitigate seasonal hydrological variances, such as flood-level water expansion, images were strictly selected from the dry season (November--March).

The curated dataset was divided into training, validation, and primary test sets. The primary test set refers to the held-out portion of the same curated dataset and was used to evaluate segmentation performance under the same data distribution. In addition, a separate geographically unseen evaluation set was prepared from riverbank regions that were not included during training, validation, or primary testing. This unseen set was used to examine how well the adapted model generalizes to out-of-domain riverbank conditions.

\begin{figure}[!t]
    \centering
    \includegraphics[width=0.77\linewidth]{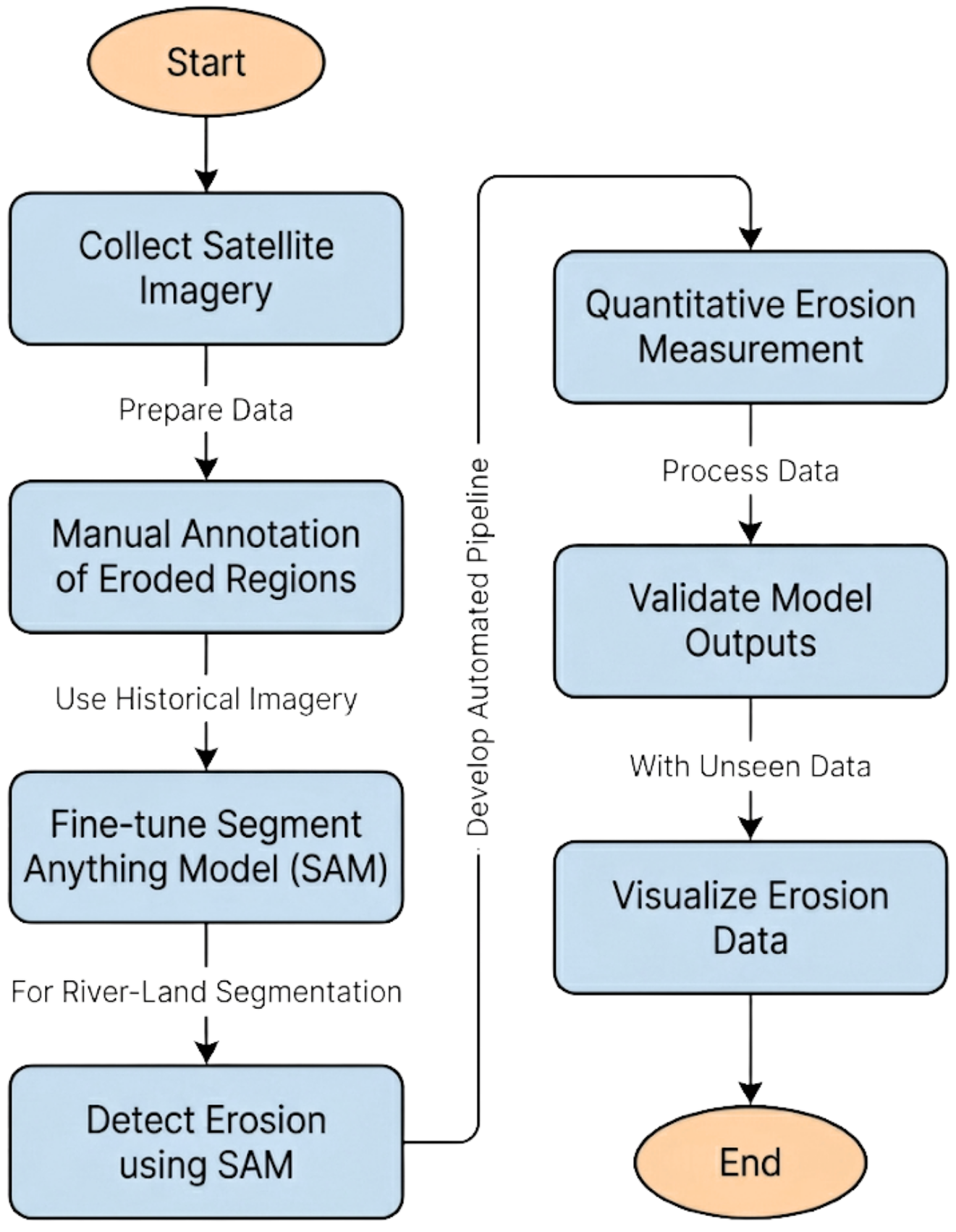}
    \caption{Overview of the proposed automated riverbank erosion monitoring pipeline. The process begins with historical satellite image collection and manual annotation of erosion-prone regions, followed by parameter-efficient fine-tuning of the Segment Anything Model (SAM) for river--land segmentation. The trained model is then used for erosion detection, quantitative land-loss estimation, validation on unseen data, and visualization of spatiotemporal erosion patterns.}
    \label{fig:erosion_pipeline}
\end{figure}

\subsubsection{Preprocessing}
Since the source imagery varied significantly in spatial resolution (0.5 m to 15 m), all samples were spatially resampled to a uniform ground sampling distance (GSD) of 10 m/pixel using bicubic interpolation. This ensures consistent feature scale for the vision transformer. Pixel intensities were normalized to the range $[0, 1]$, and histogram equalization was applied to harmonize contrast across different sensors.

\subsubsection{Annotation}
Ground truth masks were generated using the Computer Vision Annotation Tool (CVAT). Annotators manually delineated three semantic classes: \textit{River}, \textit{Stable Land}, and \textit{Eroded Area}. To ensure quality, a dual-blind review process was employed, yielding a Cohen’s Kappa inter-annotator agreement score of 0.87.

\subsection{Model Architecture and Adaptation}

We adopted the Segment Anything Model (SAM) \cite{kirillov2023sam} as the backbone architecture. SAM consists of a heavy ViT-H image encoder, a prompt encoder, and a lightweight mask decoder.


\begin{figure}[!t]
    \centering
    \includegraphics[width=\linewidth]{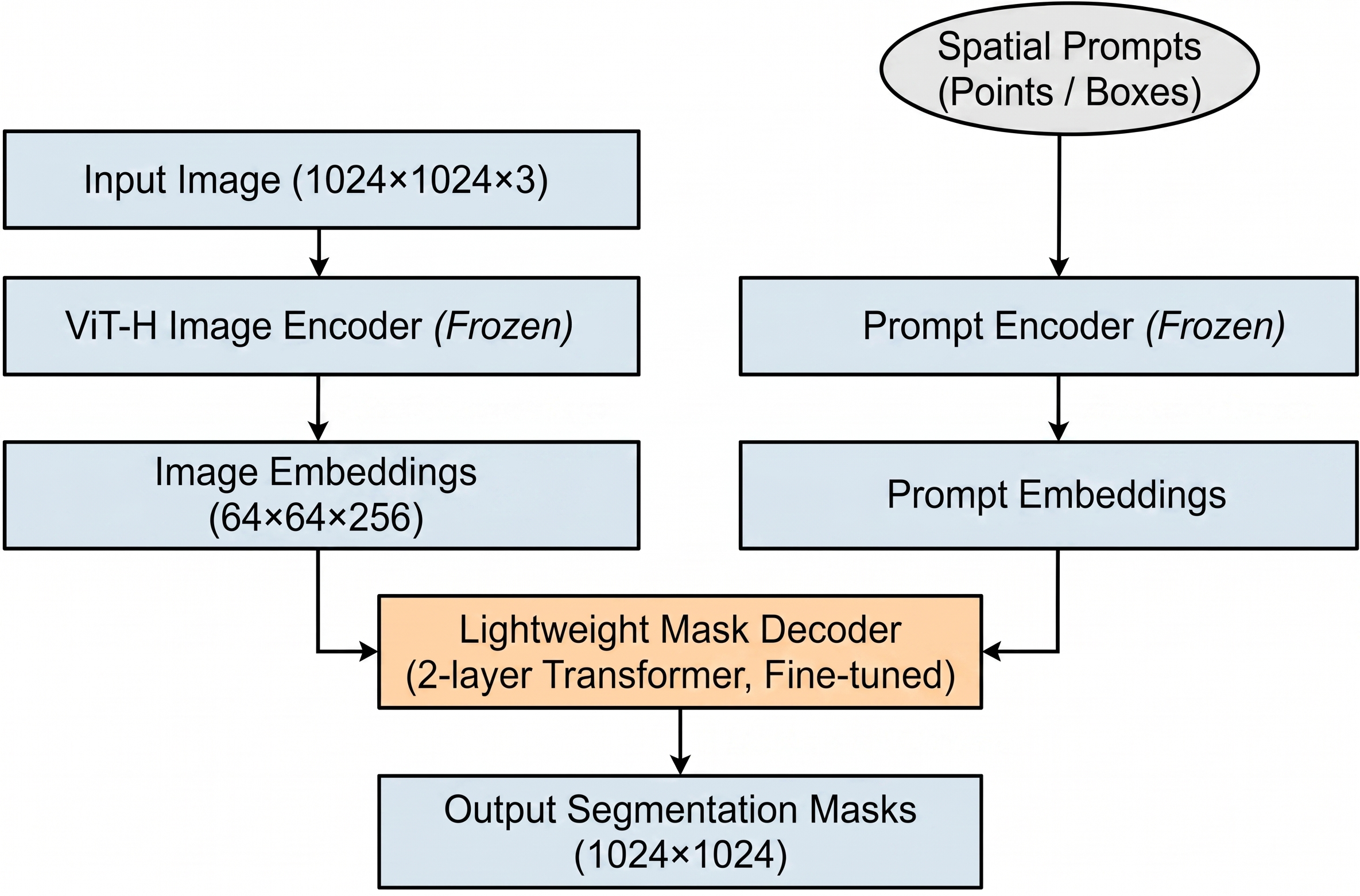}
    \caption{Architecture of the proposed parameter-efficient fine-tuned Segment Anything Model (SAM) framework for riverbank erosion segmentation. The ViT-H image encoder and prompt encoder remain frozen during training, while only the lightweight mask decoder is fine-tuned using spatial prompts to generate high-resolution segmentation masks.}
    \label{fig:sam_architecture}
\end{figure}

\subsubsection{Fine-Tuning Strategy}
Standard SAM struggles with the spectral ambiguity of sediment-laden water (which visually resembles sandy riverbanks). To adapt the model without the computational cost of retraining the ViT-H encoder (632M parameters), we employed a parameter-efficient fine-tuning (PEFT) strategy. We froze the image encoder and prompt encoder, updating only the mask decoder weights (4M parameters).

\subsubsection{Prompt Generation During Training}
To simulate user interaction and guide the model during training, we dynamically generated point prompts. For each training iteration, random foreground point prompts were sampled from the ground truth erosion masks and fed into the prompt encoder as spatial guidance signals. This forces the model to learn the specific textural features of eroded riverbanks conditioned on spatial hints.

\subsection{Training Protocol}

Training was conducted on an NVIDIA RTX 5070 Ti (16GB VRAM) using PyTorch 2.0. The optimization objective was a composite loss function designed to handle class imbalance (erosion pixels are a minority class):

\begin{equation}
L_{\text{total}} = \lambda_{foc} L_{\text{focal}} + \lambda_{dice} L_{\text{dice}} + \lambda_{iou} L_{\text{iou}}
\end{equation}

where $\lambda_{foc}=20.0$, $\lambda_{dice}=1.0$, and $\lambda_{iou}=1.0$. We used the Focal Loss ($L_{focal}$) to penalize hard-to-classify examples, defined as:
\begin{equation}
L_{\text{focal}} = -\alpha (1-p_t)^{\gamma}\log(p_t)
\end{equation}
with focusing parameter $\gamma=2.0$ and balancing factor $\alpha=0.25$. Optimization utilized AdamW with a learning rate of $1 \times 10^{-4}$ and cosine annealing decay over 50 epochs.

\subsection{Quantification of Morphological Change}

To quantify land loss, we implemented a temporal difference pipeline.  Let $M_{land}^{t}$ denote the binary segmentation mask of land predicted by SAM at time $t$. For an image pair $(t_1, t_2)$ where $t_1 < t_2$, the semantic change masks are defined as:

\begin{equation}
M_{\text{erosion}} = M_{land}^{t_1} \cap \neg M_{land}^{t_2}
\end{equation}
\begin{equation}
M_{\text{accretion}} = \neg M_{land}^{t_1} \cap M_{land}^{t_2}
\end{equation}

Post-processing includes morphological opening to remove noise (connected components $<500$ pixels). The physical area of erosion ($A_{loss}$) is calculated as:
\begin{equation}
A_{loss} = \sum_{i,j} M_{\text{erosion}}(i,j) \times r^2
\end{equation}
where $r=10$ m is the pixel resolution. This enables the direct conversion of pixel counts into hectares or square kilometers for impact assessment.

\begin{figure}[!t]
\centering

\begin{subfigure}[b]{0.48\linewidth}
    \centering
    \includegraphics[
        width=\linewidth,
        height=0.40\textheight,
        keepaspectratio
    ]{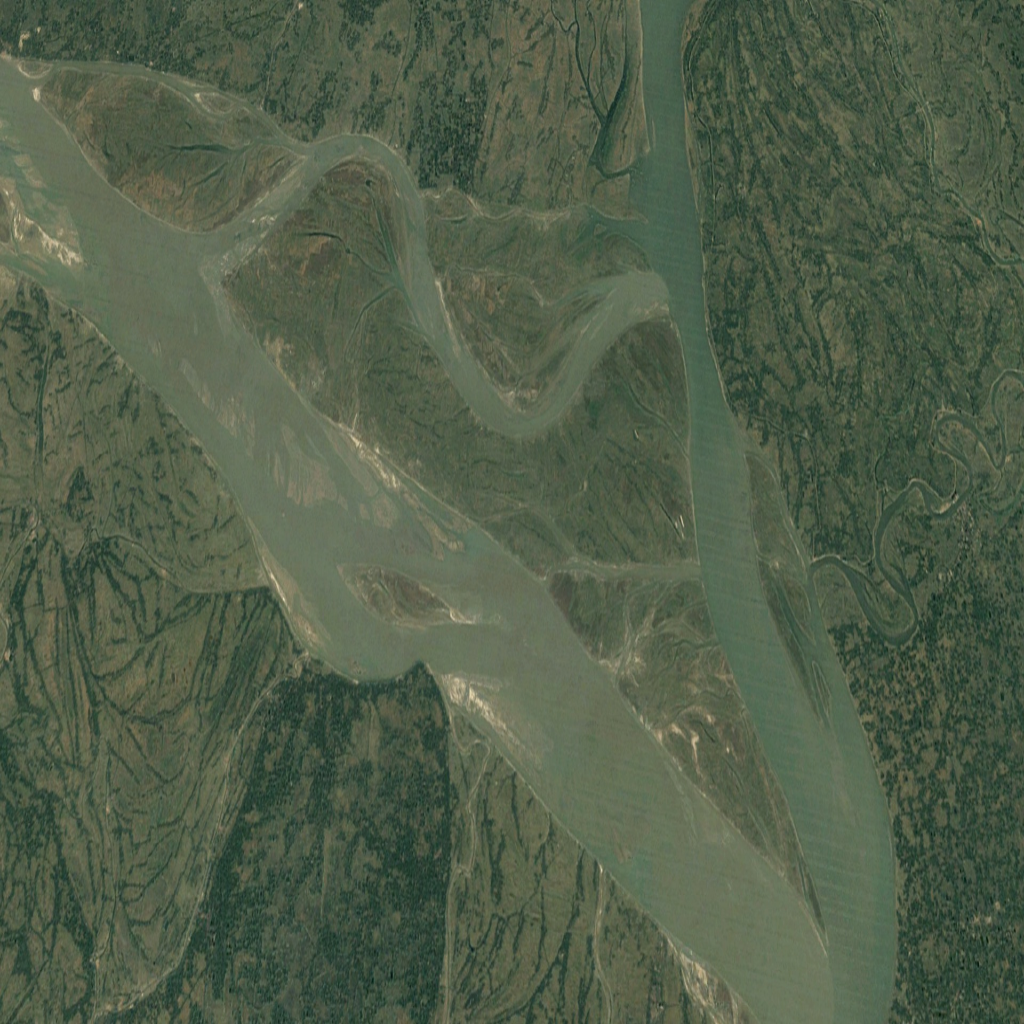}
    \caption{Satellite image of Naria Upazila (2010).}
\end{subfigure}\hfill
\begin{subfigure}[b]{0.48\linewidth}
    \centering
    \includegraphics[
        width=\linewidth,
        height=0.40\textheight,
        keepaspectratio
    ]{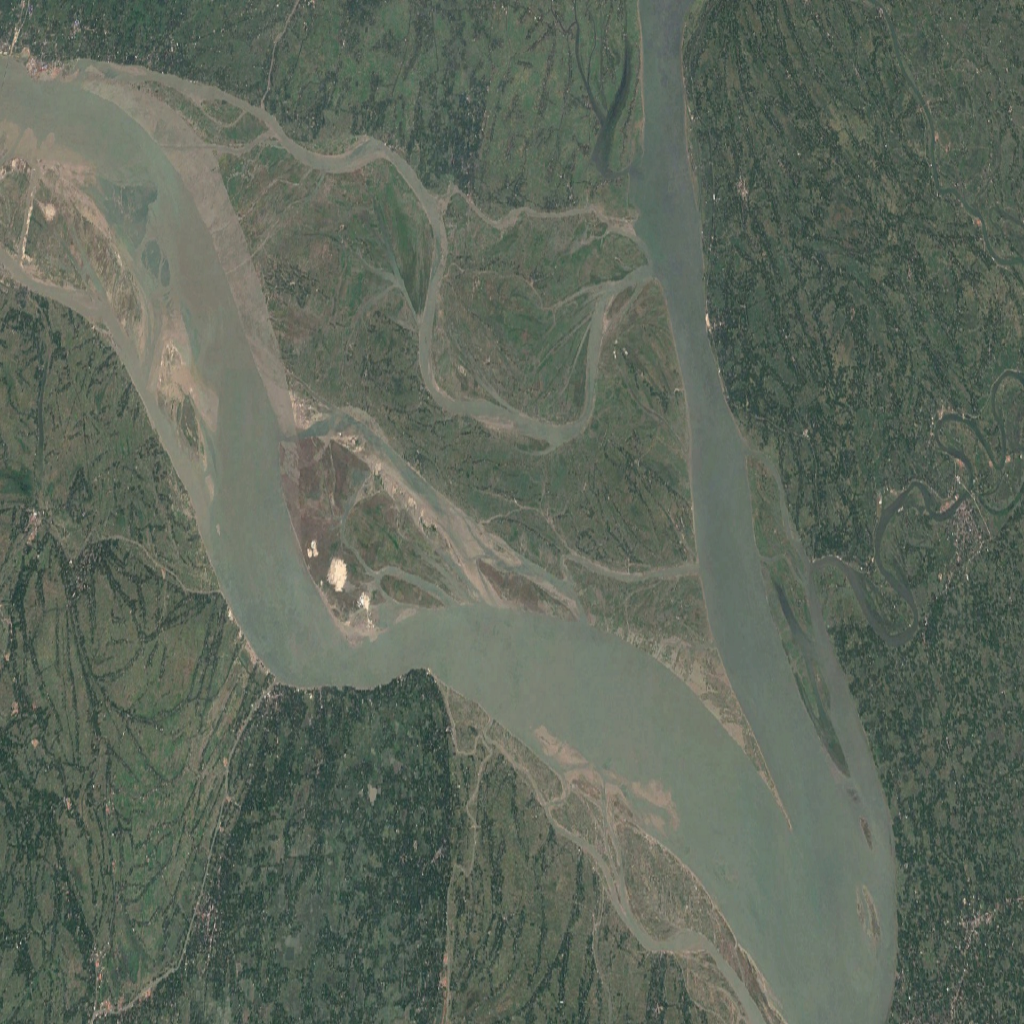}
    \caption{Satellite image of Naria Upazila (2020).}
\end{subfigure}

\vspace{0.4em}

\begin{subfigure}[b]{0.48\linewidth}
    \centering
    \includegraphics[
        width=\linewidth,
        height=0.26\textheight,
        keepaspectratio
    ]{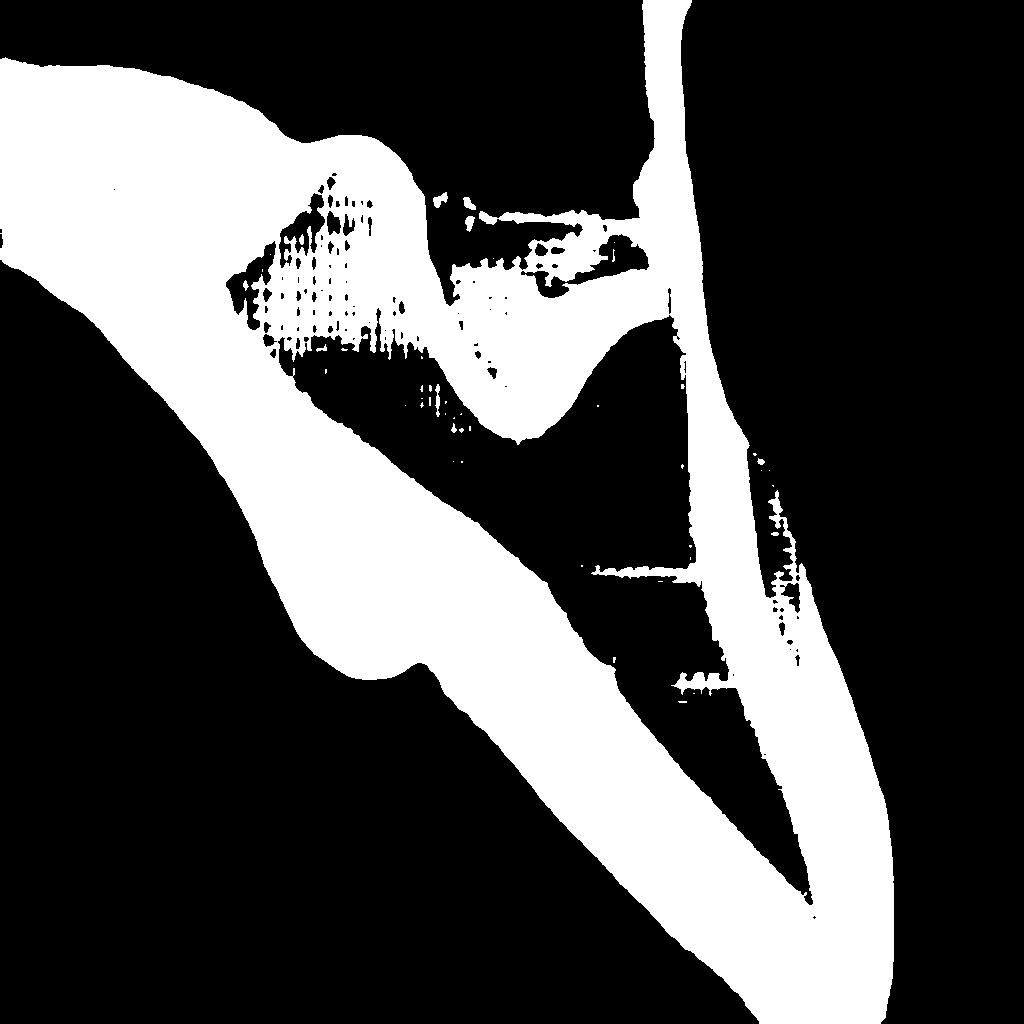}
    \caption{Base SAM (zero-shot), view 1.}
\end{subfigure}\hfill
\begin{subfigure}[b]{0.48\linewidth}
    \centering
    \includegraphics[
        width=\linewidth,
        height=0.26\textheight,
        keepaspectratio
    ]{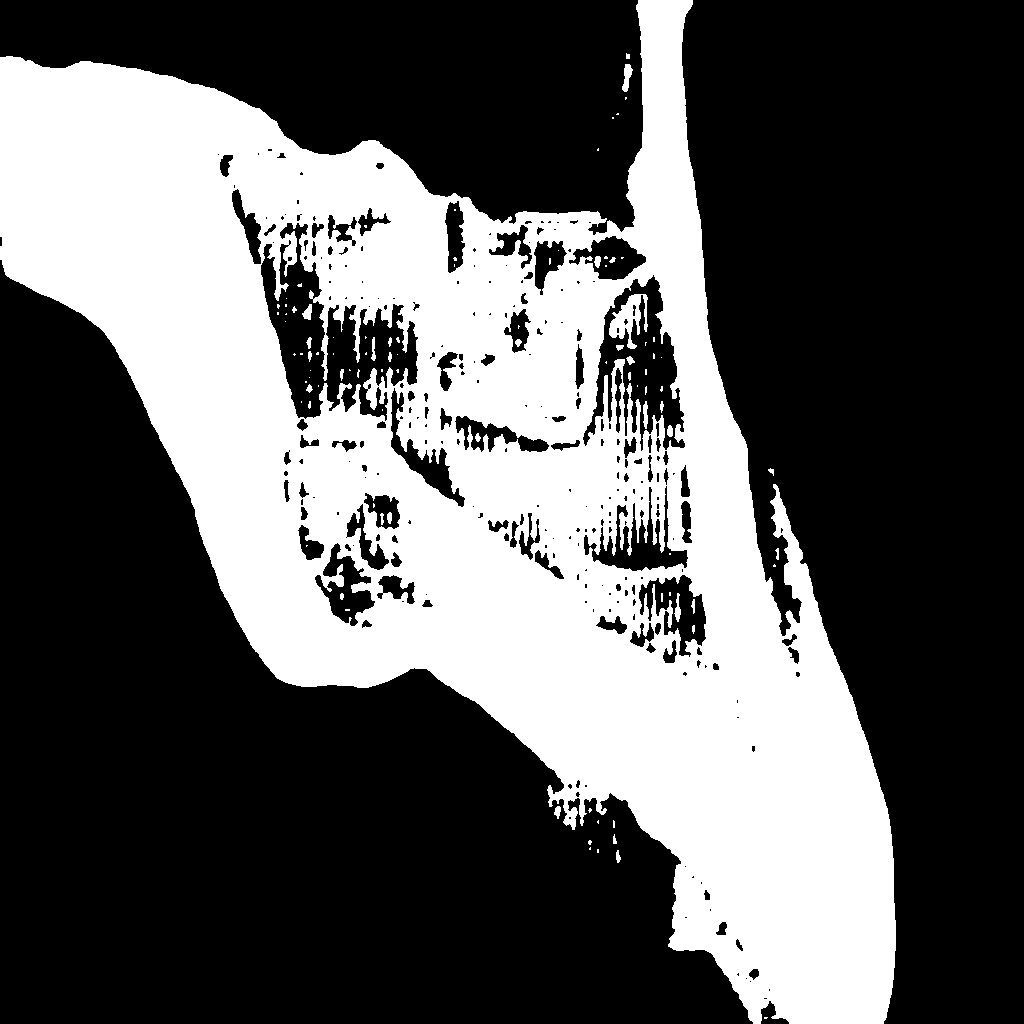}
    \caption{Base SAM (zero-shot), view 2.}
\end{subfigure}

\vspace{0.4em}

\begin{subfigure}[b]{0.48\linewidth}
    \centering
    \includegraphics[
        width=\linewidth,
        height=0.26\textheight,
        keepaspectratio
    ]{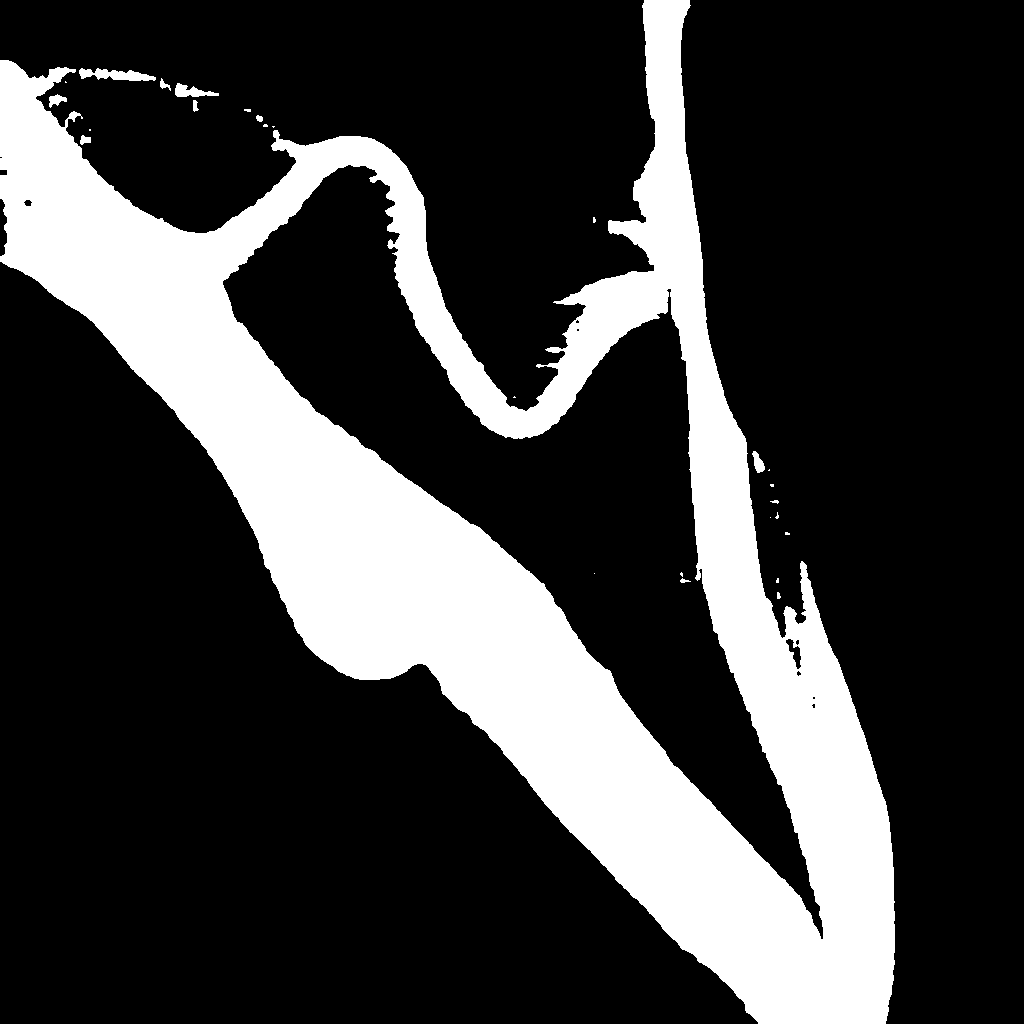}
    \caption{Fine-tuned SAM model, view 1.}
\end{subfigure}\hfill
\begin{subfigure}[b]{0.48\linewidth}
    \centering
    \includegraphics[
        width=\linewidth,
        height=0.26\textheight,
        keepaspectratio
    ]{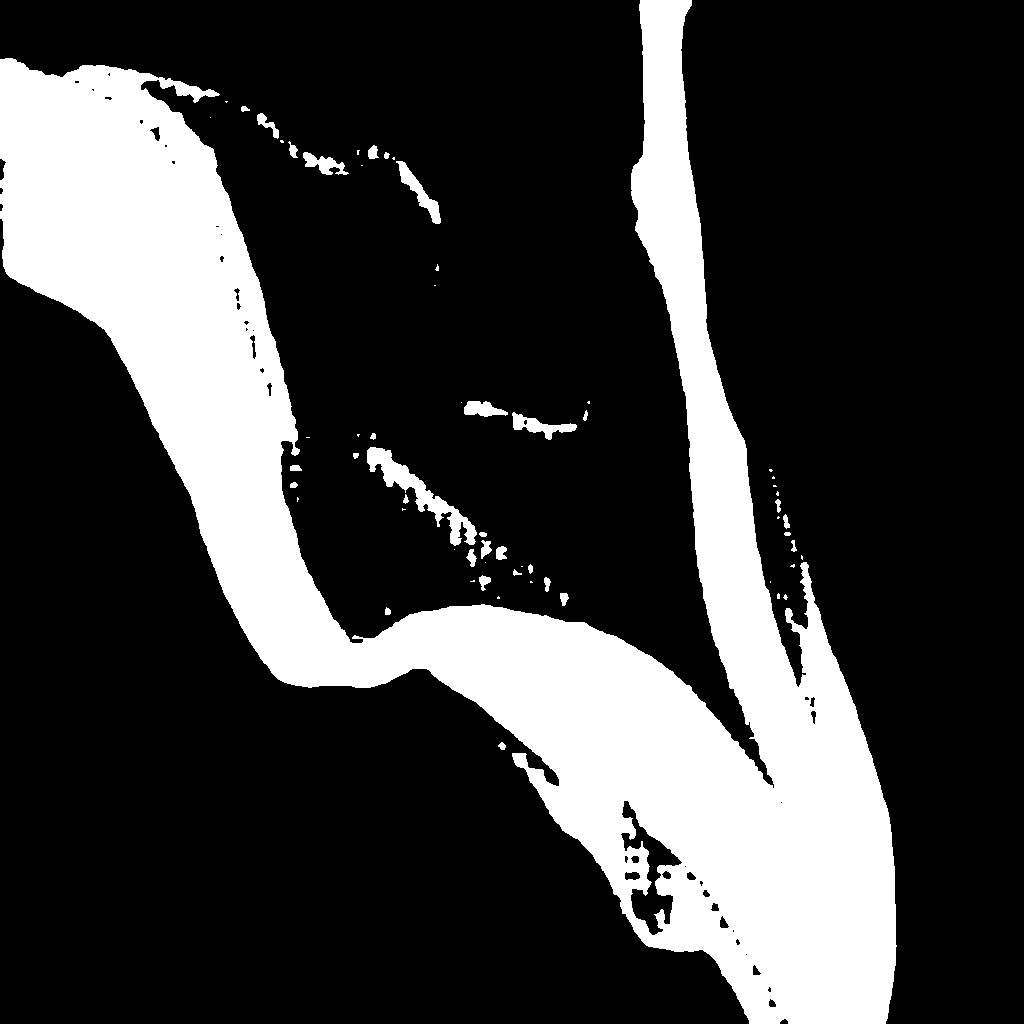}
    \caption{Fine-tuned SAM model, view 2.}
\end{subfigure}

\caption{Comparison of Naria Upazila erosion patterns and SAM-generated masks.
Top row: Satellite imagery (2010 vs.\ 2020).
Middle row: Base SAM zero-shot segmentation results.
Bottom row: Fine-tuned SAM predictions with improved delineation of erosion zones.}
\label{fig:naria_comparison}

\end{figure}

\section{Results and Discussion}

\subsection{Training Stability and Convergence}
The fine-tuning process showed stable convergence, with the training loss decreasing consistently across epochs. Early stopping was used to reduce the risk of overfitting, and the learning rate scheduler adjusted the learning rate as the validation loss plateaued.

\subsection{Segmentation Performance}

Table~\ref{tab:test_metrics} reports the overall fine-tuned model performance across evaluated classes, while Table~\ref{tab:sam_comparison} reports the erosion-class comparison between the base SAM and the fine-tuned SAM.

\begin{table}[!t]
\caption{Segmentation Performance on the Primary Held-Out Test Set}
\label{tab:sam_comparison}
\centering
\footnotesize
\begin{tabular}{l r r}
\toprule
\textbf{Metric} & \multicolumn{1}{c}{\textbf{Base SAM (Zero-Shot)}} & \multicolumn{1}{c}{\textbf{Fine-Tuned SAM}} \\
\midrule
IoU             & 0.655 & \textbf{0.867} \\
Precision       & 0.661 & \textbf{0.971} \\
Recall          & \textbf{0.988} & 0.891 \\
F1-score        & 0.789 & \textbf{0.928} \\
Pixel Accuracy  & 0.882 & \textbf{0.970} \\
\bottomrule
\end{tabular}
\end{table}

The quantitative results in Table~\ref{tab:sam_comparison} are reported on the primary held-out test set from the curated dataset. These results show the limitations of the zero-shot base SAM and the effectiveness of domain adaptation through decoder-only fine-tuning.

\subsubsection{Base SAM vs. Fine-Tuned SAM}
The Zero-Shot Base SAM exhibited a high Recall (0.988) but poor Precision (0.661). This indicates that the base model was overly aggressive, frequently misclassifying wet sand or vegetation as water bodies. By contrast, the Fine-Tuned SAM achieved a Precision of 0.971, a 47\% improvement. Although Recall dropped slightly to 0.891, the overall F1-score increased from 0.789 to 0.928, confirming that the adapted model learned to distinguish the subtle spectral signatures of true river boundaries versus temporary waterlogging.

\subsubsection{Boundary Delineation}
As shown in Table \ref{tab:test_metrics}, the model achieved a Boundary IoU of 0.825. This metric is critical for erosion monitoring, as the meaningful morphological changes occur strictly at the edges. Visual inspection (Fig. \ref{fig:naria_comparison}) confirms that while Base SAM output "blob-like" predictions, the Fine-Tuned model captured the jagged, irregular coastlines characteristic of riverbank shear failure.

\begin{table}[!h]
\centering
\caption{Overall Fine-Tuned Model Performance on the Primary Held-Out Test Set}
\label{tab:test_metrics}
\begin{tabular}{l r}
\toprule
\textbf{Metric} & \multicolumn{1}{c}{\textbf{Value}} \\
\midrule
Mean IoU & 0.801 \\
F1-Score & 0.887 \\
Precision & 0.915 \\
Recall & 0.861 \\
Boundary IoU & 0.825 \\
\bottomrule
\end{tabular}
\end{table}

\subsection{Quantification of Morphological Change}

\begin{figure}[t]
\centering
\begin{subfigure}{\linewidth}
    \centering
    \includegraphics[width=\linewidth]{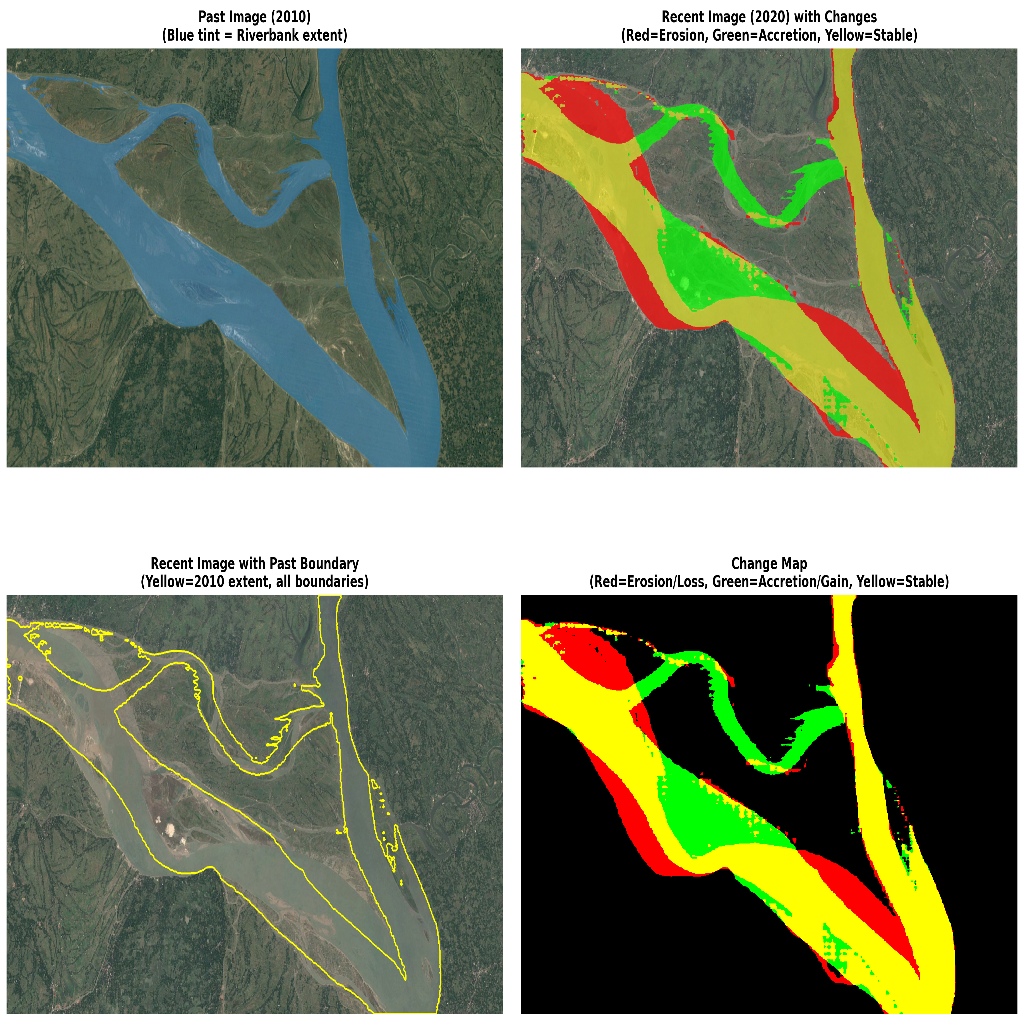}
    \caption{Ground Truth Change Analysis}
    \label{fig:riverbank_gt}
\end{subfigure}
\vspace{2mm}
\begin{subfigure}{\linewidth}
    \centering
    \includegraphics[width=\linewidth]{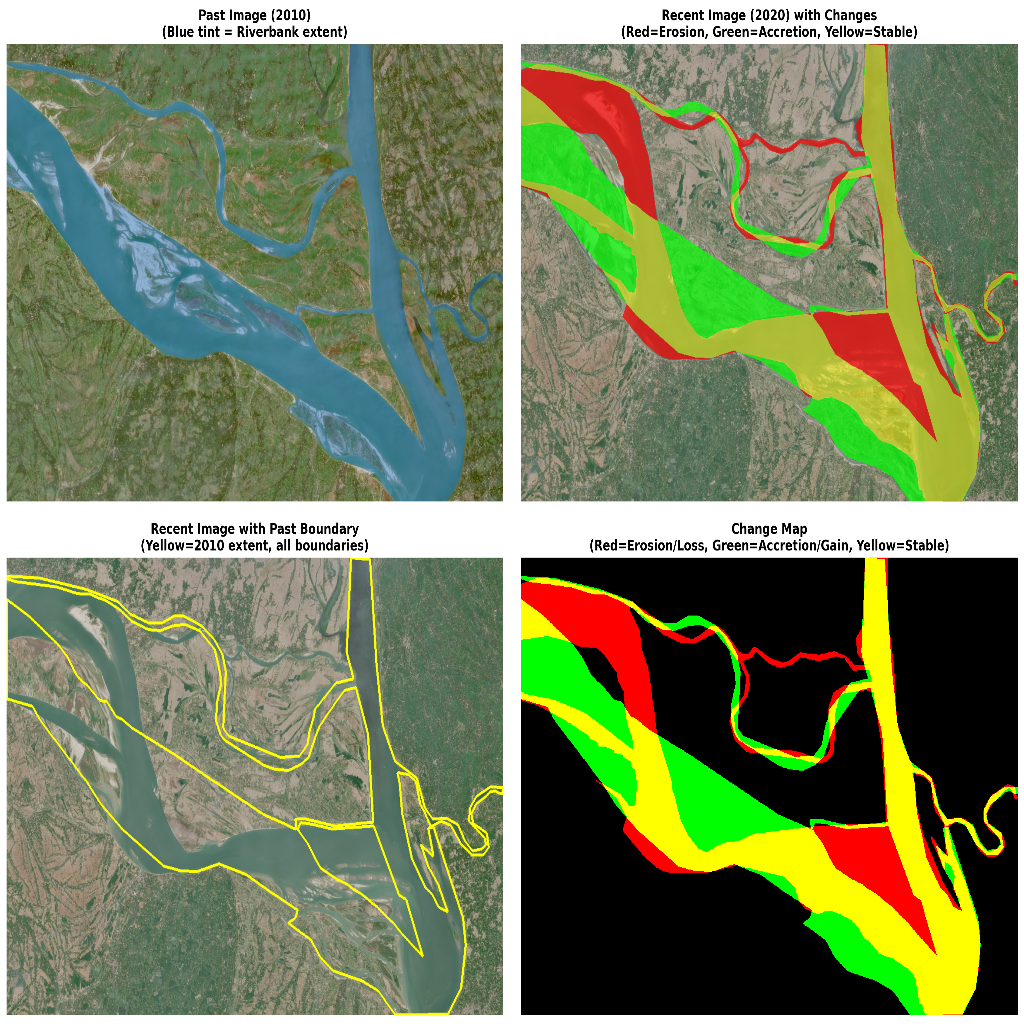}
    \caption{Predicted Change Analysis (Fine-Tuned SAM)}
    \label{fig:riverbank_sam}
\end{subfigure}
\caption{Visual comparison of riverbank change quantification. The model accurately captures the erosion zones (red) but underestimates accretion (green).}
\label{fig:riverbank_change_analysis}
\end{figure}

The ultimate utility of this framework lies in quantifying land loss. For this analysis, the model was further evaluated on a geographically unseen riverbank region to examine performance outside the primary dataset split. Table~\ref{tab:change_analysis} compares the area calculations derived from ground truth masks and model predictions on this unseen evaluation case.

\subsubsection{Erosion Detection Accuracy}
The model proved exceptionally accurate in quantifying erosion. It estimated a total land loss of \textbf{11.772 km²}, differing from the ground truth (11.752 km²) by only \textbf{0.17\%}.  This suggests that the visual cues of erosion—such as sharp cut-banks and immediate vegetation loss—are distinct features that the SAM decoder can learn robustly.

\subsubsection{The Accretion Challenge}
Conversely, the model struggled to quantify accretion (land gain), underestimating it by approximately 14.5\% (12.072 km² predicted vs. 14.118 km² actual). This underestimation resulted in an over-calculation of "Stable" land. 
\textbf{Analysis of Error:} This discrepancy is likely due to the spectral similarity between "Stable Sand" and "Newly Accreted Sand." In RGB imagery, both appear as bright, textured regions. Without multispectral bands (e.g., Infrared) or elevation data (DEM) to distinguish established soil from fresh deposits, the vision-only model tends to classify new accretion as pre-existing land.

\begin{table}[!t]
\caption{Change Analysis on the Geographically Unseen Evaluation Set (km$^2$)}
\label{tab:change_analysis}
\centering
\footnotesize
\begin{tabular}{lcc}
\toprule
\textbf{Category} & \textbf{Ground Truth} & \textbf{Fine-tuned SAM} \\
\midrule
Erosion (Land Loss)     & 11.752 & \textbf{11.772} \\
Accretion (Land Gain) & 14.118 & 12.072 \\
Stable (Unchanged) & 33.002 & 37.869 \\
\midrule
\textbf{Net Change} & \textbf{+2.365} & \textbf{+0.299} \\
\bottomrule
\end{tabular}
\end{table}

\section{Conclusion}

This study presented an automated framework for riverbank erosion monitoring in Bangladesh using a parameter-efficient adaptation of the Segment Anything Model (SAM) and a curated set of historical Google Earth image pairs. The work shows that foundation-model-based segmentation can be adapted for riverine change analysis while reducing the need for fully manual interpretation of satellite imagery. At the same time, the study has several limitations that should be addressed in future research. The current framework relies on RGB-only imagery, which limits its ability to distinguish stable sandy riverbanks from newly accreted sediment. Future work should therefore incorporate multispectral data, Near-Infrared bands, water indices such as NDWI, and elevation information to improve the separation of water, exposed sand, vegetation, and newly formed land. The present model is also based on dry-season imagery, which restricts its use for continuous monsoon-season monitoring under cloud cover and high-water conditions. Integrating Synthetic Aperture Radar (SAR) can help enable all-weather and year-round riverbank observation. In addition, future datasets should include more small-scale erosion events, accretion zones, and visually ambiguous riverbank regions to reduce class imbalance and improve generalization. Beyond segmentation and area estimation, future work may also include temporal modeling of long-term river migration and the integration of population, infrastructure, and socio-economic layers so that automated erosion monitoring can better support disaster preparedness, planning, and riverbank risk assessment.

\bibliographystyle{IEEEtran}
\bibliography{references}

\end{document}